\documentclass{article}

\PassOptionsToPackage{round}{natbib}
\usepackage[preprint,main]{neurips_2026}
\let\cite\citep

\usepackage[utf8]{inputenc}
\usepackage[T1]{fontenc}
\usepackage[hidelinks]{hyperref}
\usepackage{url}
\usepackage{graphicx}
\usepackage{booktabs}
\usepackage{amsfonts}
\usepackage{nicefrac}
\usepackage{amsmath}
\usepackage{microtype}
\usepackage{xcolor}
\usepackage{multirow}
\usepackage{placeins}
\usepackage[breakable]{tcolorbox}
\newtcolorbox{promptbox}[1][]{
  colback=gray!10,
  colframe=gray!50,
  boxrule=0.4pt,
  arc=2pt,
  left=6pt,right=6pt,top=4pt,bottom=4pt,
  fontupper=\small\ttfamily,
  breakable,
  #1
}

\newcommand\blfootnote[1]{%
  \begingroup
  \renewcommand\thefootnote{}\footnote{#1}%
  \addtocounter{footnote}{-1}%
  \endgroup
}
\title{Forecasting Future Behavior as a Learning Task}
\author{%
  Mosh Levy\\
  Bar-Ilan University\\
  Constellation\\
  \texttt{moshe0110@gmail.com}
  \And
  Yoav Goldberg\\
  Bar-Ilan University\\
  Allen Institute for AI
  \And
  Asa Cooper Stickland\\
  UK AI Security Institute
}

\begin{document}

\maketitle
\blfootnote{Code and data: \url{https://github.com/Mosh0110/behavior-forecasters}}

\begin{abstract}
Trust in an AI system is often anchored by explanations of how it works, which one then uses to forecast its behavior on new inputs.
For large reasoning models (LRMs), this conventional route is particularly difficult to follow: explanation methods for single token generations do not naturally generalize to long trajectories, and the trajectories themselves are often not faithful when read as natural language.
We propose an alternative that bypasses the explanation step: treat behavior forecasting as a learnable task and train \emph{Behavior Forecasters} that operates on a single reasoning trajectory to make the same forecasts one would typically seek from an explanation.
The forecaster's training data is obtained by querying the LRM with no human annotation, and its inference is done in a single forward pass.
We instantiate this approach on two tasks: how likely the LRM is to repeat its answer on re-runs, and how removing parts of the input changes its answer.
We evaluate this approach on both tasks across three diverse reasoning datasets and find that
trained Behavior Forecasters are more accurate than GPT-5.4 and Claude Opus~4.6 reading the same trajectories as naive readers, at a small fraction of their inference cost.
We find that fine-tuning the backbone end-to-end and initializing it from the target LRM are each necessary for strong performance.
These results show that the reasoning trajectory carries information about the LRM's future behavior that goes beyond what naive reading conveys.
\end{abstract}

\section{Introduction}
\label{sec:intro}
Relying on an AI system requires anticipating how it will behave on new inputs.
Conventionally, this anticipation is anchored by explanations of the model's computation, widely treated as a device for forecasting\footnote{We use \emph{forecasting} rather than \emph{predicting} to distinguish between the action of stating something about a model's future behavior and the operation of a neural network producing its output.} how it will behave on new inputs or in new scenarios~\cite{Hempel1948StudiesIT, Douglas2009ReintroducingPT, jacovi2021formalizing, doshi2017rigorous, lipton2018mythos}. This view is reflected in the way researchers in machine learning operationalize and evaluate explanations by the accuracy of the forecasts they enable~\cite{miller2018explanationartificialintelligenceinsights}.
The most basic granularity at which such forecasts can drive deployment decisions is the individual input, for example how consistent the model's answer is across reruns or how it would change under counterfactual variants of the input~\cite{ribeiro2016should, lundberg2017unified, sundararajan2017axiomatic, alvarezmelis2018robustness, barredo2020explainable}.
These per-input behavior forecasts are the focus of this paper (Section~\ref{sec:prediction_task}).

Explanations that enable such forecasts are particularly difficult to obtain for large reasoning models (LRMs) that emit a long sequence of reasoning tokens before producing a final answer (e.g., O1~\cite{openai2024o1}, R1~\cite{deepseek2025r1}) (Section~\ref{sec:methods_fall_short}).
Methods developed for a single forward pass tell us about the generation of one token, and these methods do not naturally extend to the long reasoning trajectories these models emit.
The reasoning tokens themselves, despite their natural-language appearance, are often not a faithful account of the computation that produced the answer (see Section~\ref{sec:reading_text} for supporting literature), so \emph{naive reading} of the trajectory may lead to wrong forecasts.
This leaves two unsatisfying options for behavior forecasting on LRMs: read the tokens naively and accept an unreliable picture, or disregard them.

In this paper, we propose an alternative to making such forecasts via explanations: train an external model to forecast future behavior directly from a single reasoning trajectory (Section~\ref{sec:method}).
We call this trained model a \emph{Behavior Forecaster}: it relies on the tokens in the reasoning trajectory to carry information about the underlying computation, but does not require this information to be recoverable by reading the tokens as text~\cite{levy2025state}.
We treat behavior forecasting as a supervised learning problem in which the training data is generated at scale without human annotation by querying the LRM itself, producing ground-truth behavioral labels.

\begin{figure}[t]
\centering
\includegraphics[width=0.93\linewidth]{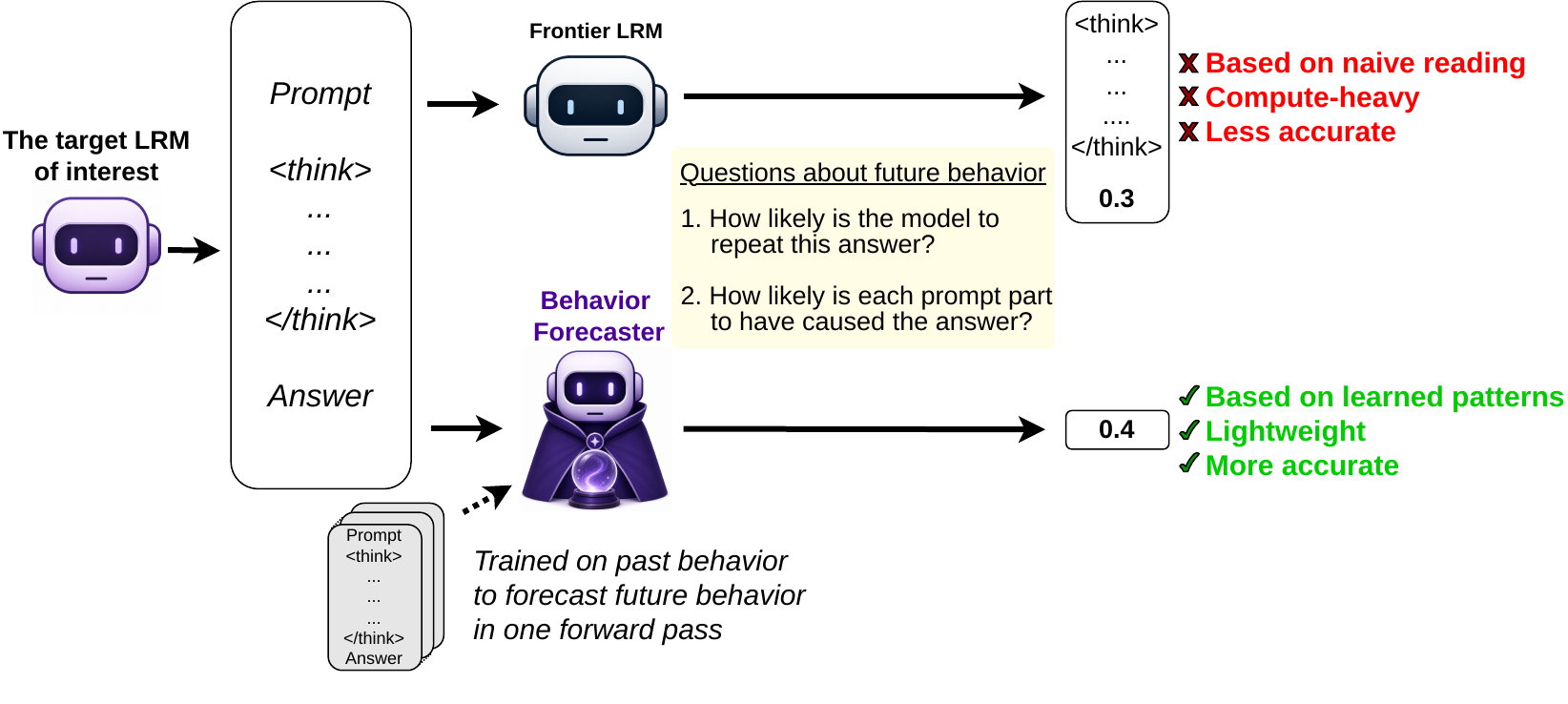}
\caption{Behavior forecasting from a single reasoning trajectory.
Given one observed trajectory of the target LRM (prompt, reasoning, answer), a trained Behavior Forecaster predicts a property of the LRM's future behavior in one forward pass: how likely the LRM is to repeat the answer (rerun consistency), or how likely each prompt part is to have caused the answer (counterfactual sensitivity).
The Behavior Forecaster is trained on many such trajectories with behavioral labels obtained by querying the LRM.
A frontier LLM reading the same trajectory naively relies on the text being faithful, does not scale as it requires a lot more compute, and yields less accurate predictions.}
\label{fig:overview}
\end{figure}

Given one reasoning trajectory of the target LRM on the input of interest, a Behavior Forecaster predicts in one forward pass a statistic of the LRM's future behavior; we instantiate it on two such statistics: \emph{rerun consistency}, computed over reruns on the same input, and \emph{counterfactual sensitivity}, computed over perturbations of the input.
We initialize the forecaster from the target LRM, attach a task-specific prediction head, and train them jointly.

We show that the Behavior Forecaster successfully learns both tasks (Section~\ref{sec:results}).
We evaluate the approach on both tasks across three reasoning datasets.
The trained Behavior Forecaster is more accurate than GPT-5.4 and Claude Opus~4.6 reading the same trajectory naively, while consuming less than $1/10{,}000$ their estimated compute.
We also show that forecasters trained on two datasets transfer to a held-out third dataset with at most $600$ fine-tuning steps, on both tasks, and that within a single dataset family they generalize to held-out variants without any fine-tuning.
The counterfactual-sensitivity forecaster also transfers to predicting whether the LRM relied on a user-provided hint, which is a common way to evaluate the faithfulness of the reasoning text~\cite{turpin2023language, chen2025reasoningmodelsdontsay, Chua2025AreDR, marioriyad2025unspokenhintsaccuracyacknowledgement}.

We then ablate the forecaster to identify which architectural and training choices matter (Section~\ref{sec:architecture_results}).
Training only the prediction head, initializing from a randomized backbone, or removing the reasoning tokens from the input all reduce performance, and the best arrangement of prompt, reasoning, and answer in the input depends on the task.

Our findings suggest that core local properties of LRM behavior, normally probed by expensive resampling or by unreliable reading of the reasoning tokens, can be forecast from a single observed trajectory.
We see this as motivation to study behavior forecasting as a learnable task in its own right.
More broadly, these findings suggest that the reasoning trajectory carries information about the LRM's future behavior beyond what naive reading conveys, inviting us to treat reasoning trajectories as data with learnable patterns rather than as natural language to be read.

\section{Behavior forecasting for LRMs}
\label{sec:problem}

In this section we motivate behavior forecasting as a route to trust in AI models, formalize the task this paper addresses, and discuss why current approaches motivate a new method for LRMs.

\subsection{Trust through behavior forecasting}
\label{sec:trust_motivation}

Trust in AI models rests on the user's ability to forecast how they will behave on future inputs~\cite{lee2004trust, hoff2015trust, jacovi2021formalizing,ZHOU2026104491}.
The concept considered as ``explanation'' is widely seen as a device for making such forecasts about a system~\cite{Hempel1948StudiesIT, Douglas2009ReintroducingPT}: knowing how a system operates enables forecasting how it will behave on new inputs or in new scenarios.
Researchers in machine learning adopt the same view, operationalizing and evaluating explanations by the accuracy of the forecasts they enable~\cite{doshi2017rigorous, miller2018explanationartificialintelligenceinsights}.

\subsection{Forecasting behavior from a single trajectory}
\label{sec:prediction_task}

We focus on forecasting properties of a model's behavior on a given input, such as how consistent the model's answer is across reruns or how sensitive its answer is to counterfactual perturbations of the input, from a single observed run~\cite{ribeiro2016should,lundberg2017unified,sundararajan2017axiomatic}.
Working at this granularity is useful in practice for two reasons.
First, per-input signals can directly drive per-query decisions such as abstention, flagging, or routing to human review, which are sometimes required from deployed AI systems~\cite{eu_ai_act_2024}.
These decisions must be made at scale, which rules out collecting multiple resamples online.
And second, larger behavioral testing, debugging, and auditing pipelines can aggregate these local measurements across inputs to characterize the system's overall behavior~\cite{ribeiro2020checklist, bhatt2020explainable}.
These pipelines run at scale too, which rules out expensive per-input measurements: the per-input cost compounds across every input the pipeline covers.

\paragraph{Formal setup.}
Let $M$ be a stochastic LRM.
On a prompt $P$, one execution of $M$ produces an observed trajectory $\tau = (P, R, A)$, where $R$ is the LRM's reasoning tokens and $A$ is the final answer.
A behavior forecasting problem specifies (i) a set of related prompts $\{P'\}$ around $P$ (which may be $P$ itself for reruns), and (ii) a target $b(M, P, A)$ defined as a statistic of $M$'s answer distribution under these related prompts, conditioned on the observed answer $A$.
The task is: given one observed trajectory $\tau \sim M(P)$, forecast $b(M, P, A)$ without running $M$ on any related prompt $P'$.

We instantiate this definition on two tasks.
\begin{itemize}
    \item \textbf{Rerun consistency:} the related prompt is $P$ itself ($P' = P$); $b(M, P, A)$ is the probability that another run produces the same answer $A$.
    \item \textbf{Counterfactual sensitivity:} for each segment $s$ in the set $S(P)$ of removable segments of $P$, the related prompt is the perturbation $P' = P_{-s}$ that removes $s$ from $P$; $b(M, P, A)$ has one component per $s$, equal to how much removing $s$ reduces the probability that $M$ produces $A$.
\end{itemize}

In both, target values lie in $[0, 1]$ and are derived from $M$'s induced answer distribution.
The task is therefore regression against probability-derived targets.

\subsection{Why we need a new approach for LRMs}
\label{sec:methods_fall_short}

There are two natural approaches to behavior forecasting on an LRM, and both run into limitations that motivate a different approach: reading the reasoning trajectory as an explanation of the LRM's computation (Section~\ref{sec:reading_text}), and applying standard behavior forecasting tools (Section~\ref{sec:standard_methods_unsuitable}).

\subsubsection{Naive reading of the reasoning trajectory is often unreliable}
\label{sec:reading_text}

A behavior forecast asks, for a given input, how confidently the LRM arrived at its answer (rerun consistency) and which input clauses it actually used (counterfactual sensitivity).

The reasoning trajectory is the natural candidate source for these outcomes in an LRM~\cite{korbak2025chainthoughtmonitorabilitynew, guan2025monitoring}.
If the trajectory's natural-language surface faithfully tracked the LRM's computation, recovering each outcome would reduce to a classical natural language processing task on that surface text: confidence detection on the trajectory for rerun consistency, and tracing how the trajectory's reasoning steps connect input segments to the final answer for counterfactual sensitivity.
We refer to this approach as \emph{naive reading}: interpreting the trajectory through generic English semantics, without learned familiarity with the patterns the target LRM uses in its reasoning tokens.
Both human readers and other LRMs read naively in this sense.
But the natural-language surface of the trajectory is often not a faithful account of the computation behind the answer.

\paragraph{Omission.}
LRMs omit factors from the natural-language reasoning that shape the final answer.
Planted cues such as biased answer positions, user hints, and stereotype signals shift predictions without appearing in the natural-language reasoning~\cite{turpin2023language, yee2024dissociation, Chua2025AreDR, arcuschin2025chain, marioriyad2025unspokenhintsaccuracyacknowledgement, lindsey2025biology, mirtaheri2026catching, young2026lietome, bachmann2026potentialcotreasoningcloser}.

\paragraph{Semantic mismatch.}
Even the steps that do appear in the natural-language reasoning can diverge from the computation the LRM does to produce the answer.
The surface natural language can be illegible to humans and AI naive readers while the LRM still arrives at correct answers~\cite{jose2025reasoningmodelsoutputillegible}, injecting changes into the natural-language reasoning often leaves the final answer unchanged~\cite{lanham2023measuring, paul2024makingreasoningmattermeasuring}, and causal mediation analysis shows that LRMs do not use their intermediate steps in the way a human reader would predict~\cite{levy2025humansperceivewrongnarratives}.

\citet{chen2025reasoningmodelsdontsay} and \citet{han2026rfeval} also show that faithfulness is not guaranteed to improve with model scale or more performant models.
Naive reading is therefore unlikely to reliably deliver the outcomes we are after, motivating a learned approach to behavior forecasting.

\subsubsection{Standard behavior forecasting methods are not applicable}
\label{sec:standard_methods_unsuitable}

Standard behavior forecasting methods are unlikely to be useful for forecasting LRMs behavior either.
The long, stochastic trajectories LRMs produce make these methods either prohibitively expensive or blind to the trajectory itself.

\paragraph{Resampling methods are prohibitively expensive.}
Resampling underlies existing methods for forecasting future model behavior~\cite{jones2025forecasting, serrano2026frontier}. It is straightforward in principle but impractical in deployment. Estimating either rerun consistency from $N$ reruns or counterfactual sensitivity from $N$ perturbations multiplies the already high cost of one long reasoning run by $N$.

\paragraph{Single-location probes miss the trajectory.}
Standard attribution and uncertainty signals all explain the generation of a single fixed position in the model's output.
Gradient-based input attributions like integrated gradients~\cite{sundararajan2017axiomatic} require a differentiable path from input to output, but the LRM computational graph involves discrete sampling steps that break the gradient flow; in LLMs the common substitute is the attention from the final-answer tokens back to the prompt~\cite{chuang2024lookbacklens, jain2019attention}.
Token-probability uncertainty signals~\cite{manakul2023selfcheckgpt, farquhar2024semanticentropy} score the final-answer logits in the same way.
None of these registers what happens inside the reasoning trajectory, where much of the LRM's computation actually unfolds.

\section{Method}
\label{sec:method}

Section~\ref{sec:problem} showed that standard tools for behavior forecasting either break down or become prohibitively expensive on LRMs, and that naive reading of the reasoning trajectory is unlikely to faithfully answer behavior forecasting questions.
We propose an alternative: we train a model to forecast behavioral properties of the LRM from a single observed reasoning trajectory.
We call this trained model a \emph{Behavior Forecaster} (or simply \emph{forecaster}).
We first describe the learning setup (Section~\ref{sec:learning_problem}), then the per-task supervision (Section~\ref{sec:task_supervision}), and finally the Behavior Forecaster architecture (Section~\ref{sec:bf_architecture}).

\subsection{Behavior forecasting as a learning task}
\label{sec:learning_problem}

We treat behavior forecasting as a supervised learning task: from one observed reasoning trajectory, predict a target behavioral statistic of the LRM.
This relies on the trajectory carrying information about the underlying computation, but does not require this information to be readable as natural-language text~\cite{levy2025state}.
Framing it this way lets behavior forecasting benefit from data and compute~\cite{sutton2019bitter}.

To generate training data, for each prompt $P$ we first obtain one observed trajectory $\tau \sim M(P)$ with answer $A$.
We then run the LRM on the related executions defined by the forecasting problem and use those executions to estimate the target statistic $b(M, P, A)$.

We train a Behavior Forecaster to map the observed trajectory to the target statistic in inference.
The expensive executions are needed only for data generation.
The cost gap relative to resampling at inference is large: estimating the rerun-consistency label for one input from a fresh resample requires additional trajectories from the LRM, each up to thousands of tokens generated autoregressively, while the Behavior Forecaster needs one forward pass over the already-observed trajectory, orders of magnitude less compute and far lower latency at deployment.

\subsection{Task-specific supervision}
\label{sec:task_supervision}

We estimate the answer probabilities entering $b(M, P, A)$ empirically: for each prompt $P$ we run the LRM $10$ times on $P$, and for counterfactual sensitivity we additionally run the LRM $10$ times on each perturbed version of $P$, all with the LRM's recommended sampling settings (temperature $0.6$, top-$p$ $0.95$).
We extract a final answer from each trajectory and treat two runs as producing the same answer when their extracted answers match, rather than requiring identical output text.
A sample is labeled only if at least $5$ of its runs yields an extractable answer; full details are in Appendix~\ref{sec:appendix_task_details}.

\paragraph{Rerun consistency.}
For each run on a prompt, the run-level target is the fraction of the prompt's other valid runs that produce the same extracted answer; this estimates the chance that a fresh run on the same prompt would reproduce the answer.

\paragraph{Counterfactual sensitivity.}
For each prompt $P$, we generate one perturbed prompt $P_{-s}$ per segment $s \in S(P)$; the segmentation scheme is dataset-specific.
We retain only prompts whose answer $A$ appears in at least $70\%$ of the $10$ original runs, so labels reflect omission effects rather than base sampling noise.
The segment-level label measures how much removing $s$ reduces the rate at which the LRM produces $A$, normalized to $[0, 1]$ so a value of $1$ means removing $s$ always changes the answer.
Every token in segment $s$ shares this score; the per-segment prediction aggregates the forecaster's per-token outputs within the segment.

\subsection{Behavior Forecaster architecture}
\label{sec:bf_architecture}

The Behavior Forecaster's backbone shares its architecture with the target LRM and initializes from the LRM's weights, so it starts with representations already attuned to the reasoning tokens.
We use the same kind of backbone for both tasks, training a separate copy per task and pairing each with a task-specific input arrangement and head.

\paragraph{Counterfactual sensitivity: prompt-echo arrangement.}
For counterfactual sensitivity, the task is one score per segment of the prompt; we aggregate per-token outputs into a per-segment score (Appendix~\ref{sec:appendix_training_objectives}).
The natural placement for these per-token predictions is at the prompt positions.
But the backbone is a standard autoregressive decoder: a prediction head at a prompt-position token attends only to earlier tokens, which excludes the reasoning $R$ and the answer $A$.
Without those, the head has none of the trajectory's evidence about how the LRM used each prompt segment.
We therefore append a copy of the prompt after the observed trajectory and produce per-token predictions at the echoed-prompt positions.
A per-token MLP head produces one prediction at each echoed-prompt position, and the per-segment score in $[0, 1]$ is obtained by aggregating these within each segment.

\paragraph{Rerun consistency: cross-attention pooling.}
For rerun consistency, the target is one score per trajectory, with no prompt-side structure to attribute to.
Trajectories vary in length, so we pool the backbone's hidden states into a fixed-size representation that a simple MLP head can map to one scalar in $[0, 1]$.
We pool with cross-attention over a fixed set of learned query vectors; the queries attend over all trajectory positions, so the pooled representation can draw on evidence from anywhere in the trajectory.
Architectural details (number of query vectors, head sizes, etc.) are in Appendix~\ref{sec:appendix_arch_and_transfer}.
    
\section{Behavior Forecasting is learnable}
\label{sec:results}

\begin{figure}[!htbp]
\centering
\includegraphics[width=\linewidth]{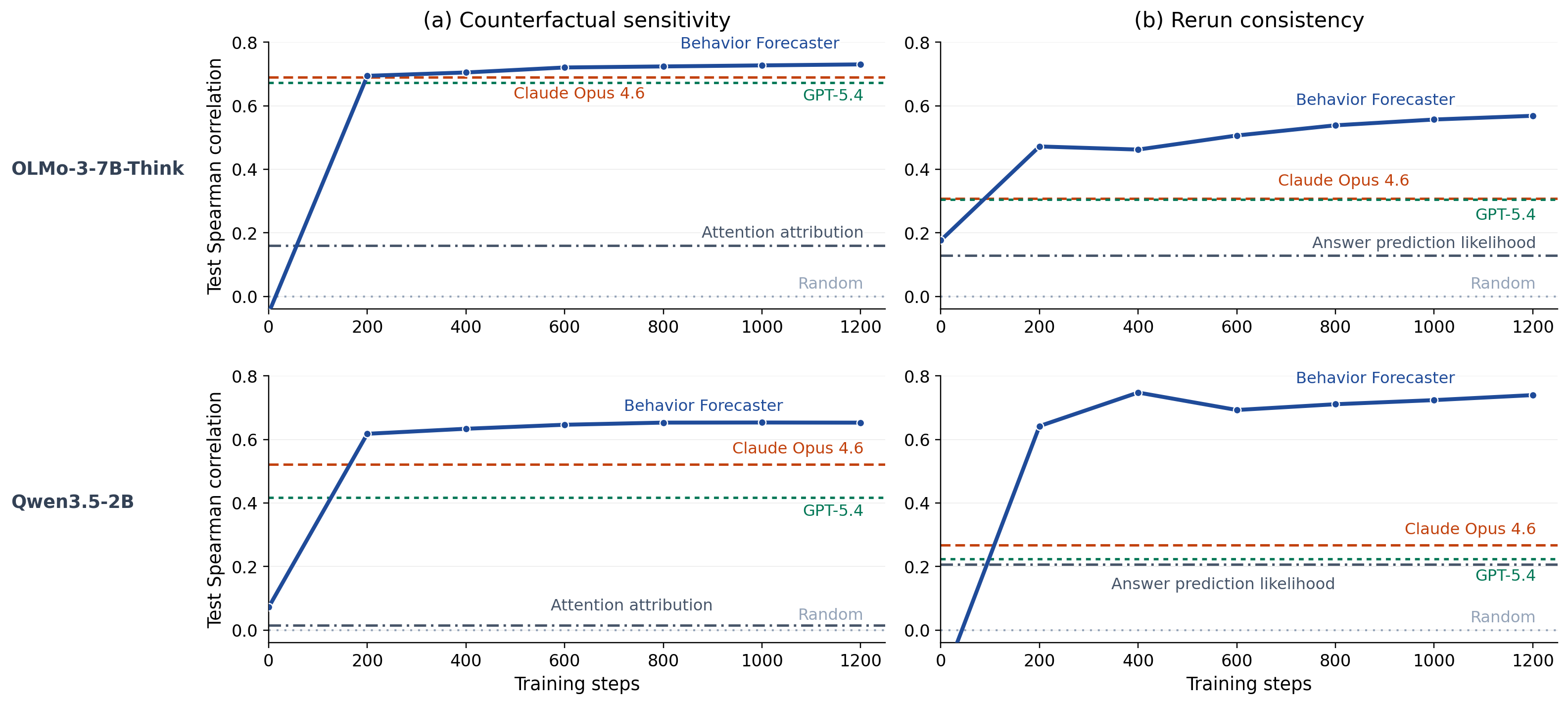}
\caption{A trained Behavior Forecaster is more accurate than two frontier naive readers and the standard single-location probes, at a small fraction of the readers' inference cost.
Rows show target LRM: OLMo-3-7B-Think (top; FEVEROUS, RuleTaker, TreeCut) and Qwen3.5-2B (bottom; FEVEROUS, RuleTaker).
For Qwen3.5-2B, TreeCut is omitted because its trajectories were too long for our available generation resources.
Each panel plots the forecaster's test Spearman correlation over training steps; horizontal lines mark the two frontier naive readers (GPT-5.4, Claude Opus~4.6), the task-specific single-location probe on the target LRM (attention attribution for counterfactual sensitivity, answer prediction likelihood for rerun consistency), and a random-prediction baseline.
Final Spearman and Pearson values for the OLMo Behavior Forecaster and both naive readers are reported in Table~\ref{tab:appendix_naive_reader_metrics}; per-dataset OLMo values, with Behavior Forecaster entries taken from the last saved checkpoint, are reported in Table~\ref{tab:appendix_main_per_dataset}; cluster-bootstrap confidence intervals and paired permutation tests for the OLMo comparison are reported in Appendix~\ref{sec:appendix_statistical_uncertainty}; single-location probe details are in Appendix~\ref{sec:appendix_single_location_probes}.}
\label{fig:learnability_overview}
\end{figure}

We evaluate Behavior Forecasters on both tasks across three reasoning datasets.
A Behavior Forecaster trained on the pooled three datasets is more accurate than both strong naive readers and the standard single-location probes, at a small fraction of the readers' inference cost (Section~\ref{sec:learnability}); the same comparison holds on a second target LRM, Qwen3.5-2B.
It also generalizes beyond its training data (Section~\ref{sec:transfer}): within a single dataset family the forecaster generalizes zero-shot to a held-out variant, and across datasets a forecaster trained on two of the three adapts to the held-out third with target-side fine-tuning.
Finally, we test whether a forecaster trained for counterfactual sensitivity can initialize a new Behavior Forecaster for a related hint-sensitivity target (Section~\ref{sec:hint_sensitivity_transfer}).

\subsection{Experimental setup}
\label{sec:setup}

\paragraph{Target LRMs and datasets.}
Our main target LRM is OLMo-3-7B-Think~\cite{olmo3}, whose training data and training pipeline are fully released; this lets us build evaluation sets from data we can verify the LRM was not trained on, and lets follow-up work that depends on knowing the LRM's training data or pipeline build directly on the same target.
For the main learnability comparison (Section~\ref{sec:learnability}) we additionally evaluate on Qwen3.5-2B~\cite{qwen35_2026} as a second target LRM, to check that the gap against naive readers is not specific to OLMo; all other experiments use OLMo only.
We use three reasoning datasets that cover substantively different tasks: mathematical word problems (TreeCut~\cite{treecut}), factual claims grounded in Wikipedia (FEVEROUS~\cite{feverous}), and synthetic logical reasoning (RuleTaker~\cite{clark2020ruletaker}).
Full data construction and splits details, and per-dataset sample counts are in Appendix~\ref{sec:appendix_datasets_splits}.

\paragraph{Training.}
We initialize the Behavior Forecaster from the target LRM's weights and train it end-to-end with AdamW, weight decay $0.01$, and a linear-decay schedule with $100$ warmup steps and a minimum learning rate of $1{\times}10^{-6}$.
We use an effective batch size of $32$ and train for $2$ epochs.
Counterfactual sensitivity uses learning rate $1{\times}10^{-5}$ and a token-level MLP head; rerun consistency uses learning rate $5{\times}10^{-5}$ and a sequence-level cross-attention MLP head.
All the Behavior Forecasters in the paper were trained on $4$ NVIDIA H200 GPUs in under $24$ hours.
Architectural hyperparameters and the transfer fine-tuning budget for Table~\ref{tab:transfer_main} are in Appendix~\ref{sec:appendix_arch_and_transfer}.
All results in the paper, except for the main learnability figure (Figure~\ref{fig:learnability_overview}), are evaluated on a checkpoint selected by validation loss; the main figure plots the test correlation across training steps directly.

\paragraph{Naive reader baselines.}
We compare against frontier LRMs (Claude Opus~4.6 and GPT-5.4) used as strong naive readers that receive the reasoning trajectory as input and predict the target property from its surface meaning.
These readers come well-equipped for what the reading task reduces to, but they have not learned how the target LRM uses its reasoning tokens.
\citet{maslej2025artificialintelligenceindexreport} report that LRMs perform at or above human level on the kind of basic text-understanding tasks naive reading reduces to.
In practice, researchers already use frontier LLMs as readers of reasoning trajectories, deploying them to judge whether a target LRM's trajectory tracks its underlying computation~\cite{chen2025reasoningmodelsdontsay, young2026lietome}.
The Behavior Forecaster and naive readers differ in their exposure to the target LRM's reasoning trajectories: the Behavior Forecaster is trained on those trajectories and so can learn the patterns the LRM uses to encode information in its reasoning tokens.
These readers are larger than the Behavior Forecaster and use their own extended reasoning, making them substantially more expensive to run; any margin a Behavior Forecaster shows against them is therefore a conservative measure of the trained-versus-naive gap.
For each task we tested three prompt variants on a small pilot slice and selected, per reader, the variant with the strongest pilot correlation.
Full prompts and selected variants are in Appendix~\ref{sec:appendix_reader_prompts}.

\paragraph{Metric.}
We report Spearman correlation in the main text, which depends only on the rank order of predictions and is insensitive to compression, shift, or any monotone rescaling of the predicted scale.
Additional results using Pearson correlation are in the appendix.

\subsection{Trained Behavior Forecasters are more accurate than untrained baselines}
\label{sec:learnability}

\paragraph{Naive readers.}
The trained Behavior Forecaster is more accurate than strong naive readers (GPT-5.4 and Claude Opus~4.6) reading the same trajectories, at a small fraction of their inference cost, on both target LRMs (Figure~\ref{fig:learnability_overview}).
On Qwen3.5-2B the Behavior Forecaster reaches Spearman $0.653$ on counterfactual sensitivity (vs $0.417$ for GPT-5.4 and $0.522$ for Claude Opus~4.6) and $0.740$ on rerun consistency (vs $0.224$ and $0.267$ respectively).
We also find that a same-LRM reader control, in which OLMo-3-7B-Think reads its own observed trajectory, is substantially weaker than both the trained Behavior Forecaster and the frontier naive readers (Appendix~\ref{sec:appendix_target_model_naive_reader}).

\paragraph{Single-location probes.}
The Behavior Forecaster also outperforms the single-location probes from Section~\ref{sec:standard_methods_unsuitable} applied to the target LRM: final-answer attention to candidate input segments for counterfactual sensitivity~\citep{chuang2024lookbacklens,jain2019attention} and final-answer log-likelihood for rerun consistency~\citep{manakul2023selfcheckgpt,farquhar2024semanticentropy}.
Both clear the random-prediction baseline but remain too weak to be useful in practice (on the OLMo target, $0.160$ and $0.129$ respectively; on Qwen3.5-2B, $0.016$ and $0.206$), well below the Behavior Forecaster on both targets and below the naive readers (Appendix~\ref{sec:appendix_single_location_probes}).

\subsection{Generalization to new data}
\label{sec:transfer}

We test how a trained Behavior Forecaster generalizes to data it was not trained on, in two settings.

\paragraph{Same-family check.}
We train a Behavior Forecaster on three subsets of the FEVEROUS dataset, holding out the numerical subset entirely, and then evaluate it directly on FEVEROUS numerical examples with no fine-tuning.
The Behavior Forecaster remains ahead of both frontier naive readers on this held-out subset for both tasks (Table~\ref{tab:feverous_near_ood}).

\paragraph{Transfer across datasets.}
We hold out one of our three datasets (FEVEROUS, RuleTaker, or TreeCut), train a source Behavior Forecaster on the remaining two, and fine-tune it on the held-out target for at most ${\sim}600$ steps.
Transfer improves performance on every held-out target for both tasks, though rerun consistency remains weaker on TreeCut than on FEVEROUS or RuleTaker after adaptation (Table~\ref{tab:transfer_main}).

\begin{table}[!htbp]
\centering
\small
\caption{Same-family transfer within FEVEROUS, for the OLMo-3-7B-Think Behavior Forecaster.
The Behavior Forecaster is trained on the non-numerical FEVEROUS subsets and evaluated on FEVEROUS numerical without target-side fine-tuning.}
\label{tab:feverous_near_ood}
\setlength{\tabcolsep}{8pt}
\begin{tabular}{l cc cc}
\toprule
& \multicolumn{2}{c}{\textit{Counterfactual sensitivity}} & \multicolumn{2}{c}{\textit{Rerun consistency}} \\
\cmidrule(lr){2-3} \cmidrule(lr){4-5}
\textbf{System} & Spearman & Pearson & Spearman & Pearson \\
\midrule
Behavior Forecaster & $\mathbf{0.565}$ & $\mathbf{0.730}$ & $\mathbf{0.311}$ & $\mathbf{0.350}$ \\
GPT-5.4              & $0.322$          & $0.406$          & $0.223$          & $0.226$          \\
Claude Opus~4.6      & $0.463$          & $0.622$          & $0.173$          & $0.193$          \\
\bottomrule
\end{tabular}
\end{table}

\begin{table}[!htbp]
\centering
\small
\caption{Transfer to held-out datasets, for the OLMo-3-7B-Think Behavior Forecaster.
The \emph{source} Behavior Forecaster is trained on the two other datasets and evaluated on the held-out target without target-side training; the \emph{adapted} Behavior Forecaster is the fine-tuned checkpoint.
Src and Adp denote source and adapted.}
\label{tab:transfer_main}
\setlength{\tabcolsep}{6pt}
\begin{tabular}{l cccc cccc}
\toprule
& \multicolumn{4}{c}{\textit{Counterfactual sensitivity}} & \multicolumn{4}{c}{\textit{Rerun consistency}} \\
\cmidrule(lr){2-5} \cmidrule(lr){6-9}
& \multicolumn{2}{c}{Spearman} & \multicolumn{2}{c}{Pearson} & \multicolumn{2}{c}{Spearman} & \multicolumn{2}{c}{Pearson} \\
\cmidrule(lr){2-3} \cmidrule(lr){4-5} \cmidrule(lr){6-7} \cmidrule(lr){8-9}
\textbf{Held-out target} & Src & Adp & Src & Adp & Src & Adp & Src & Adp \\
\midrule
FEVEROUS  & $0.368$ & $0.641$ & $0.319$ & $0.757$ & $0.242$ & $0.539$ & $0.233$ & $0.471$ \\
RuleTaker & $0.446$ & $0.661$ & $0.692$ & $0.962$ & $0.005$ & $0.438$ & $0.238$ & $0.854$ \\
TreeCut   & $0.650$ & $0.846$ & $0.584$ & $0.938$ & $0.118$ & $0.193$ & $0.157$ & $0.295$ \\
\bottomrule
\end{tabular}
\end{table}

\subsection{Transfer across behavioral targets}
\label{sec:hint_sensitivity_transfer}
We also test transfer to a new behavioral target.
Starting from the counterfactual-sensitivity Behavior Forecaster, we adapt to hint-sensitivity forecasting: given a single trajectory produced with an explicit answer hint, predict the probability that the target LRM's no-hint answer would differ from the observed hinted answer.
The test split holds out two strong hint templates and contains $610$ trajectories from $122$ FEVEROUS source examples.
As shown in Table~\ref{tab:dir6_hint_transfer}, the transferred forecaster obtains the best rank correlation on this surface, outperforming both direct hint-sensitivity training from the target LRM and frontier naive readers that see the same hinted prompt, observed answer, and full completion.

\begin{table}[!htbp]
\centering
\small
\caption{Transfer to hint-sensitivity forecasting on held-out FEVEROUS hint templates, for the OLMo-3-7B-Think Behavior Forecaster.
The direct Behavior Forecaster is trained on the hint-sensitivity surface from the target LRM initialization; the transferred Behavior Forecaster initializes from a counterfactual-sensitivity model and is then adapted to the same hint-sensitivity surface.
All methods are evaluated on the same $610$ held-out trajectories.}
\label{tab:dir6_hint_transfer}
\setlength{\tabcolsep}{10pt}
\begin{tabular}{lcc}
\toprule
\textbf{System} & \textbf{Spearman} & \textbf{Pearson} \\
\midrule
Behavior Forecaster (transferred) & $\mathbf{0.800}$ & $\mathbf{0.817}$ \\
Behavior Forecaster (direct)      & $0.732$          & $0.814$          \\
Claude Opus~4.6 reader            & $0.768$          & $0.779$          \\
GPT-5.4 reader                    & $0.668$          & $0.680$          \\
\bottomrule
\end{tabular}
\end{table}

\section{Ablating the Behavior Forecaster}
\label{sec:architecture_results}

We now ask which parts of the Behavior Forecaster make its learning possible: the input arrangement, the pretrained initialization, and end-to-end backbone training.
We vary the input arrangement (which of prompt $P$, reasoning $R$, and answer $A$ are present and in what order), whether the backbone is initialized from the target LRM or randomly initialized with the same architecture, and whether the backbone is frozen while only the prediction head is trained.
Table~\ref{tab:sec5_ablation_results} reports the resulting test correlations.

\begin{table}[!htbp]
\centering
\footnotesize
\caption{Behavior Forecaster variants across input arrangement, initialization, and backbone training.
Test Spearman and Pearson for OLMo-backed ablations.
$P$, $R$, and $A$ denote prompt, reasoning tokens, and final answer.
``OLMo init'' is short for OLMo-3-7B-Think initialization.
The bold row in each task is the main Behavior Forecaster used in Section~\ref{sec:results}.}
\label{tab:sec5_ablation_results}
\setlength{\tabcolsep}{5pt}
\begin{tabular}{lcc lcc}
\toprule
\multicolumn{3}{c}{\textit{Counterfactual sensitivity}} & \multicolumn{3}{c}{\textit{Rerun consistency}} \\
\cmidrule(lr){1-3} \cmidrule(lr){4-6}
\textbf{Variant} & Spearman & Pearson & \textbf{Variant} & Spearman & Pearson \\
\midrule
OLMo init, $P$--$R$--$A$--$P$ & $\mathbf{0.723}$ & $\mathbf{0.925}$ & OLMo init, $P$--$R$--$A$ & $\mathbf{0.568}$ & $\mathbf{0.649}$ \\
OLMo init, $P$--$A$--$P$ & $0.714$ & $0.909$ & OLMo init, $P$--$R$--$A$--$P$ & $0.554$ & $0.653$ \\
OLMo init, $P$--$R$--$A$ & $0.578$ & $0.620$ & OLMo init, $P$--$A$ & $0.316$ & $0.251$ \\
OLMo init, prompt only & $0.567$ & $0.602$ & OLMo init, prompt only & $0.348$ & $0.279$ \\
OLMo arch., random init & $0.534$ & $0.553$ & OLMo arch., random init & $0.477$ & $0.504$ \\
Frozen OLMo, head-only & $0.245$ & $0.251$ & Frozen OLMo, head-only & $0.475$ & $0.552$ \\
Random predictions & $0.000$ & $0.000$ & Random predictions & $0.000$ & $0.000$ \\
\bottomrule
\end{tabular}
\end{table}

\paragraph{The useful arrangement is task-specific.}
Counterfactual sensitivity is strongest with the prompt-echo $P$--$R$--$A$--$P$ arrangement, while rerun consistency is strongest with the plain $P$--$R$--$A$ arrangement; swapping one task's preferred arrangement for the other's hurts performance on both.
Removing the reasoning tokens hurts both tasks substantially, so the reasoning tokens themselves, not just the prompt and answer, carry the signal the Behavior Forecaster uses.

\paragraph{Initialization from the target LRM helps.}
On both tasks, initializing from OLMo-3-7B-Think is stronger than a randomly initialized backbone with the same architecture.
The gap is larger for counterfactual sensitivity than for rerun consistency.
This suggests that pretrained representations contribute beyond what the architecture alone provides.

\paragraph{An external trained model beats attaching only a head to the LRM.}
Attaching only a prediction head to the original LRM, i.e., freezing the OLMo backbone and training only the head, performs well below the fully fine-tuned Behavior Forecaster (a separately trained copy of the backbone with its own head).
The original LRM's hidden states already contain useful signal, but adapting a separate model end-to-end on top of those states substantially improves extraction of that signal.

\section{Related work}
\label{sec:related_work}

\paragraph{Probing internal states for current-run properties.}
A line of work trains probes on the internal states of language models to predict properties of the model's behavior on the current input.
Such methods predict whether a generated statement is factual~\cite{azaria2023internal, burns2022discovering}, whether the model will hallucinate~\cite{kossen2024semanticentropyprobes, kadavath2022know}, whether the model will refuse a request~\cite{arditi2024refusal}, and broader behavioral attributes such as honesty or power-seeking tendencies~\cite{zou2025representationengineeringtopdownapproach}; earlier probing work also recovers behaviorally relevant signals from the residual stream more broadly~\cite{alain2016understanding, belrose2023tuned, karvonen2025activationoracles}.
A growing line of work applies the same idea inside LRM trajectories: probes on hidden states at intermediate steps predict whether the current intermediate answer is correct, supporting early exit~\citep{zhang2025reasoningmodelsknow}; attention probes decode the same run's eventual final answer well before it appears in the generated text~\citep{boppana2026reasoningtheater}; and mechanistic analysis finds that in-context concepts are progressively refined across the trajectory, with steering experiments confirming that this refinement causally drives accuracy~\citep{kharlapenko2026fluidrepresentations}.
All of these targets concern the current run: properties of the trajectory or its eventual answer.
Our work targets the LRM's future behavior instead: rerun consistency and counterfactual sensitivity describe the distribution over future executions related to the input, not properties of the realized one.

\paragraph{Resampling-based analysis of LRM trajectories.}
Prior work estimates the causal effect of individual reasoning steps on the answer distribution by repeatedly resampling and intervening on the trajectory~\citep{bogdan2025thoughtanchorsllmreasoning, macar2025thoughtbranches}.
For deployment-time per-query decisions, the per-input cost of these methods is prohibitive: dozens to hundreds of fresh trajectories, each thousands of forward passes long.
We treat the resampling cost as a one-time investment to label training data, and train a Behavior Forecaster that approximates the same statistics from a single observed trajectory.

\section{Limitations}
\label{sec:limitations}

\paragraph{OOD tasks generalization.}
We test transfer across datasets from the same behavioral tasks in our study, but we do not establish how well Behavior Forecasters generalize to OOD tasks that differ substantially from those used for training.
A stronger test would require training on a much broader pool of tasks.
This is costly because each labeled example requires many additional target-LRM generations, and those generations can be long and compute-intensive.
As a result, our findings provide a promising first indication that this approach can work, but deployment would require broader training coverage.

\paragraph{More faithful future LRMs could close the gap.}
\citet{swaroop2025fritcausalimportance} and \citet{baker2025monitoringreasoningobfuscation} propose explicit methods for training LRMs to produce more faithful chains of thought.
A more faithful target LRM could close part of the gap between a trained Behavior Forecaster and naive reading, since more of the relevant computation would already be visible at the surface. 

\section{Conclusion}
\label{sec:conclusion}

Trust in an AI system rests on the ability to forecast how it will behave, which is particularly challenging for LRMs (Section~\ref{sec:methods_fall_short}).
We formulated behavior forecasting as its own learning task and introduced \textit{Behavior Forecasters}, models that forecast behavioral properties of the target LRM from a single observed reasoning trajectory.
We trained Behavior Forecasters on rerun consistency and counterfactual sensitivity, with labels generated by querying the LRM itself.

The task is learnable: a trained Behavior Forecaster outperforms strong naive readers and single-location probes, generalizes across datasets with fine-tuning and across same-family variants without it, and relies on initialization from the target LRM, end-to-end fine-tuning, and a task-specific input arrangement.

Our results show that the trajectory holds more information about the LRM's future behavior than on surface level.
We hope this motivates further study of behavior forecasting as a learning task: any LRM behavioral property that can be automatically labeled is a candidate target for low-cost forecasting at deployment scale.

\bibliographystyle{plainnat}
\bibliography{bib}
\appendix
\section{Data and label construction}
\label{sec:appendix_task_details}

\subsection{Datasets and splits}
\label{sec:appendix_datasets_splits}

The paper-facing experiments use three reasoning datasets: FEVEROUS~\cite{feverous}, RuleTaker~\cite{clark2020ruletaker}, and TreeCut~\cite{treecut}.
FEVEROUS includes multiple subsets; we take its four largest labeled subsets and combine them into a single dataset for splitting and balancing.

We balance by dataset so that no single dataset dominates training and so that headline metrics are not driven by the largest available pool.
For counterfactual sensitivity, the per-dataset cap is $2{,}000$ original samples where available; RuleTaker contributes its full filtered pool of $1{,}628$ samples.
For rerun consistency, each dataset contributes $1{,}500$ original samples.
The combined FEVEROUS budget is split across its four subsets in proportion to the available pool of each subset, which preserves the natural relative size of each subset within the FEVEROUS budget.

\paragraph{Trajectory length.}
We cap the length of the target LRM's generation (reasoning tokens plus final answer) at sampling time.
We chose this cap to be long enough that the great majority of trajectories complete naturally and short enough that each item fits within the per-item memory budget of our training hardware.
We use $8{,}000$ tokens.
Across the paper-facing datasets, between $98.6\%$ and $99.8\%$ of generated runs reach a final answer within the cap.

\paragraph{Splits.}
The main learnability runs use a sample-level $95/0/5$ train/test split with seed $42$.
Runs that require validation-based checkpoint selection use an $85/10/5$ train/val/test split derived from the same held-out test surface.

\paragraph{Counterfactual sensitivity counts.}
For each original sample we sample the target LRM $10$ times on the original prompt and $10$ times on each perturbed prompt.
A sample is kept only if it has at least $5$ valid runs of the original prompt and an original-answer consistency of at least $0.7$, so its labels are built on a sufficiently stable original answer.
Each kept sample carries multiple segment-level labels, one per kept removable segment.

Table~\ref{tab:appendix_dir3_split_counts} reports the per-dataset sample counts.

\begin{table}[t]
\centering
\small
\caption{Counterfactual sensitivity sample counts for the $85/10/5$ split.
FEVEROUS is treated as one dataset, with its budget distributed across its four subsets.}
\label{tab:appendix_dir3_split_counts}
\setlength{\tabcolsep}{4pt}
\begin{tabular}{l r r r r}
\toprule
\textbf{Dataset} & \textbf{Selected samples} & \textbf{Train} & \textbf{Val} & \textbf{Test} \\
\midrule
FEVEROUS  & 2{,}000 & 1{,}700 & 200 & 100 \\
RuleTaker & 1{,}628 & 1{,}384 & 163 &  81 \\
TreeCut   & 2{,}000 & 1{,}700 & 200 & 100 \\
\midrule
Total     & 5{,}628 & 4{,}784 & 563 & 281 \\
\bottomrule
\end{tabular}
\end{table}

\paragraph{Metric aggregation.}
For counterfactual sensitivity, each retained segment label is one evaluation item.
We compute Spearman and Pearson correlations over the flattened vector of segment-level predictions and labels, rather than first averaging predictions within each original sample.
Consequently, samples with more retained segments contribute more evaluation items to the reported correlation.
Table~\ref{tab:appendix_dir3_metric_units} reports the corresponding held-out sample counts and segment-level evaluation entries.

\begin{table}[t]
\centering
\small
\caption{Evaluation units for counterfactual sensitivity correlations.
Each segment entry contributes one prediction--label pair to Spearman and Pearson.}
\label{tab:appendix_dir3_metric_units}
\setlength{\tabcolsep}{5pt}
\begin{tabular}{l r r}
\toprule
\textbf{Dataset} & \textbf{Test samples} & \textbf{Segment entries} \\
\midrule
FEVEROUS entity & 4 & 70 \\
FEVEROUS multi-hop & 39 & 1{,}105 \\
FEVEROUS numerical & 19 & 780 \\
FEVEROUS tables/text & 38 & 1{,}200 \\
RuleTaker & 81 & 11{,}740 \\
TreeCut & 100 & 1{,}425 \\
\midrule
Total & 281 & 16{,}320 \\
\bottomrule
\end{tabular}
\end{table}

\paragraph{Rerun consistency counts.}
For each original sample we sample the target LRM $10$ times on the prompt.
We require at least $5$ valid runs for a sample to be labeled, so each label is built from a sufficient sample size.
At training and evaluation time, the Behavior Forecaster consumes $5$ observed runs per sample as separate inputs.
This separates the runs used to estimate each label (up to $10$, post-filter) from the runs actually fed to the Behavior Forecaster ($5$, the inference-time constant we hold fixed across samples).

Table~\ref{tab:appendix_dir4_split_counts} reports per-dataset sample counts.

\begin{table}[t]
\centering
\small
\caption{Rerun consistency per-dataset sample counts in the $85/10/5$ split.}
\label{tab:appendix_dir4_split_counts}
\setlength{\tabcolsep}{6pt}
\begin{tabular}{l r r r r}
\toprule
\textbf{Dataset} & \textbf{Train} & \textbf{Val} & \textbf{Test} & \textbf{Total} \\
\midrule
FEVEROUS  & 1{,}275 &  150 &  75 & 1{,}500 \\
RuleTaker & 1{,}275 &  150 &  75 & 1{,}500 \\
TreeCut   & 1{,}275 &  150 &  75 & 1{,}500 \\
\midrule
Total     & 3{,}825 &  450 & 225 & 4{,}500 \\
\bottomrule
\end{tabular}
\end{table}

\subsection{Answer extraction}
\label{sec:appendix_answer_extraction}

The label-construction rules in the rest of this appendix treat two runs as producing the same answer when their dataset-specific extracted values are equal, not when their full output strings match.
We apply the parser to the LRM's post-\texttt{</think>} answer span only.
For each dataset we apply a short ordered list of regular expressions and use the first match; if none matches, the run is marked extraction-failed and excluded from the labeled run set.

\paragraph{FEVEROUS (multiple choice).}
The three FEVEROUS verdicts (\textsc{Supports}, \textsc{Refutes}, \textsc{Not enough info}) are presented as a multiple-choice question with letters A--C.
The parser looks, in order, for: (i)~a boxed letter, e.g.\ \texttt{\textbackslash boxed\{A\}}; (ii)~an explicit ``\texttt{(final\;)?answer/choice (is|:) X}'' phrasing; (iii)~a single letter on its own line; (iv)~a trailing letter at end of text.
The extracted value is the matched letter.

\paragraph{RuleTaker (true/false).}
The parser looks, in order, for: (i)~an explicit ``\texttt{label: true|false}'' phrasing; (ii)~the last occurrence of a standalone \texttt{true} or \texttt{false} token in the answer span.
The extracted value is the matched token, lowercased.

\paragraph{TreeCut (numeric or unknown).}
The parser first checks for an explicit \textsc{Unknown} answer using ``\texttt{(final\;)?answer (is|:) unknown}'' or a standalone \texttt{unknown} at the end of the answer span.
Otherwise it looks for, in order: (i)~a boxed numeric value, e.g.\ \texttt{\textbackslash boxed\{16.5\}}; (ii)~an ``\texttt{answer/result (is|:) $n$}'' phrasing; (iii)~a number at end of text; and as a last resort the last numeric literal anywhere in the answer span.
Two numeric values are treated as the same when they are exactly equal under \texttt{float} comparison; \textsc{Unknown} is its own equivalence class.

\subsection{Counterfactual sensitivity label construction}
\label{sec:appendix_dir3_labels}

For each original prompt $P$ we define a dataset-specific set of removable segments $S(P)$ and create one perturbed prompt $P_{-s}$ for each $s \in S(P)$.
A run is valid if it has no generation failure, has an extractable answer, reaches the closing \texttt{</think>} tag, and has non-empty post-\texttt{</think>} answer text.
For each kept prompt, we identify the answer $A$ produced by at least $0.7$ of the $10$ valid original runs; prompts without such an answer are dropped.

Let $c_{\mathrm{orig}}$ be the fraction of valid original runs producing $A$, and $c_{-s}$ the fraction of valid perturbed runs on $P_{-s}$ producing $A$.
The segment label is
\[
\mathrm{AF}(s) =
\mathrm{clip}\!\left(
\frac{c_{\mathrm{orig}} - c_{-s}}{c_{\mathrm{orig}}},\,0,\,1
\right).
\]
A perturbed instance is kept only if it has at least $5$ valid runs.
A sample can survive with only a subset of its segment labels; missing segments contribute no supervision during training.
This per-segment retention rule is intentional: it lets us keep a sample's well-supported segments without forcing every segment to meet the run-count threshold.

\paragraph{Omission units.}
Table~\ref{tab:appendix_dir3_segments} lists the dataset-specific unit removed to form one perturbed instance.

\begin{table}[t]
\centering
\small
\caption{Omission units used to construct counterfactual sensitivity perturbations.}
\label{tab:appendix_dir3_segments}
\begin{tabular}{l l}
\toprule
\textbf{Dataset} & \textbf{Removed unit} \\
\midrule
FEVEROUS & One evidence bullet \\
RuleTaker & One statement \\
TreeCut & One evidence paragraph \\
\bottomrule
\end{tabular}
\end{table}

\subsection{Rerun consistency label construction}
\label{sec:appendix_dir4_labels}

A run is valid if it has no generation failure and has an extractable answer.
If run $i$ has extracted answer $a_i$, and $c_i$ of the $n$ valid runs on the same input have extracted answer $a_i$, the run-level target is
\[
\ell_{\mathrm{cons}}(i) = \frac{c_i - 1}{n - 1}.
\]
This estimates the likelihood that another run on the same input will produce the same answer as run $i$.

If any of a sample's runs hits the trajectory length cap, we discard the entire sample rather than dropping individual long runs.
Discarding the whole sample preserves the exact run set its labels were computed on, so we do not need to recompute labels after the length filter.

\section{Behavior Forecaster training and architecture}
\label{sec:appendix_bf_training}

\subsection{Training objectives}
\label{sec:appendix_training_objectives}

\paragraph{Counterfactual sensitivity.}
Counterfactual sensitivity uses the dataset-defined set of removable segments $S(P)$ during training.
For an observed trajectory $\tau=(P,R,A)$ in the prompt-echo arrangement, the Behavior Forecaster emits one logit $z_t$ at each token position in the echoed prompt region.
For each retained segment $s \in S(P)$, we build a binary mask $m_{s,t}$ over the echo-region positions corresponding to that segment.
The Behavior Forecaster first pools logits within the segment,
\[
\bar{z}_s =
\frac{\sum_t m_{s,t} z_t}{\sum_t m_{s,t}},
\qquad
\hat{a}_s = \sigma(\bar{z}_s).
\]
The target $a_s$ is the clipped attributable-fraction label from Section~\ref{sec:appendix_dir3_labels}.
The loss is binary cross-entropy on the pooled segment logit,
\[
\mathcal{L}_{\mathrm{cf}}
= \frac{1}{|\mathcal{S}_{\mathrm{valid}}|}
\sum_{s \in \mathcal{S}_{\mathrm{valid}}}
\left[
-a_s \log \sigma(\bar{z}_s)
-(1-a_s)\log(1-\sigma(\bar{z}_s))
\right].
\]
Thus each retained segment contributes one loss term, even though the Behavior Forecaster produces token-level logits.
This objective assumes the removable segments and prompt-alignment masks are provided by the dataset construction; the Behavior Forecaster is trained to score the supplied segments, not to discover the segment boundaries.
At evaluation time we use the same pooled predictions and compute correlations over the flattened set of retained segment labels.

\paragraph{Rerun consistency.}
Rerun consistency is a trajectory-level prediction problem and has no segment masks.
For each observed trajectory $i$, the Behavior Forecaster emits one scalar logit $z_i$ and prediction $\hat{\ell}_i=\sigma(z_i)$.
We train with mean-squared error against the continuous run-level consistency label:
\[
\mathcal{L}_{\mathrm{rerun}}
= \frac{1}{N}\sum_{i=1}^{N}
\left(\hat{\ell}_i - \ell_{\mathrm{cons}}(i)\right)^2.
\]
Each of the $5$ observed runs per sample is treated as a separate training and evaluation item.
All rerun-consistency results reported in the paper use this continuous objective.

\subsection{Architectural hyperparameters and transfer fine-tuning budget}
\label{sec:appendix_arch_and_transfer}

\paragraph{Backbone and heads.}
The OLMo-3-7B-Think backbone has hidden size $d=4096$.
The token-level head used for counterfactual sensitivity is a $2$-layer MLP with hidden sizes $[512, 256]$ applied at each echoed-prompt position, producing one logit per token.
The sequence-level head used for rerun consistency pools the backbone's hidden states with cross-attention over $12$ learned query vectors of dimension $d$ ($32$ attention heads, dropout $0.1$), then concatenates the pooled vectors and passes them through a $2$-layer MLP with hidden sizes $[1024, 256]$ and GELU activations between layers (each layer is Linear $\to$ LayerNorm $\to$ GELU $\to$ Dropout) to produce one scalar logit.

\paragraph{Transfer fine-tuning budget for Table~\ref{tab:transfer_main}.}
For each held-out target dataset, we initialize from the source Behavior Forecaster (trained on the other two datasets with the same recipe as Section~\ref{sec:results}) and fine-tune the entire model on the target's train split with the same optimizer, weight decay, schedule, effective batch size, and per-task learning rate as the main runs.
Counterfactual sensitivity uses the $95/0/5$ surface and the final checkpoint after $2$ epochs of target-side fine-tuning; rerun consistency uses an $85/10/5$ surface for $2$ epochs and selects by best validation loss.

\section{Baselines and probes}
\label{sec:appendix_baselines}

\subsection{Naive reader prompts}
\label{sec:appendix_reader_prompts}

Each reader is given three pieces of information per evaluation example: the original input the target LRM saw, the target LRM's observed final answer for that run, and the target LRM's full completion text for that run.
The reader returns a continuous score in $[0, 1]$ and is not trained or calibrated on the task.

For each task we tested three prompt variants on a $100$-sample pilot slice of the evaluation set and selected, per reader model, the variant with the highest pilot Spearman correlation.
This per-reader selection is deliberate: it gives each reader its strongest configuration before we compare against the trained Behavior Forecaster.

\paragraph{Counterfactual sensitivity prompts.}
All three variants ask the reader to score, for each candidate segment, how strongly removing that segment would shift the target LRM's final answer away from the observed answer.
The variants differ only in framing wording.
GPT-5.4 used variant v2 (Box~\ref{box:cfvtwo}); Claude Opus~4.6 used variant v3 (Box~\ref{box:cfvthree}).
Variant v1 frames the score as ``how strongly removing the segment would reduce the chance of the same final answer'' and is included in the released prompt set.

\begin{promptbox}[title={Counterfactual sensitivity, variant v3 (Claude Opus~4.6)}, label=box:cfvthree]
Below is a problem and one observed run produced by an AI model. The problem is partitioned into candidate segments.

You will see the original problem, the full completion from that run, the observed final answer, and the candidate segments.

Your task is to estimate, for each candidate segment, how strongly that segment supports the observed answer: if you removed the segment, how likely is the model to move away from the observed answer?

Technical details: target model and sampling settings (temperature=0.6, top\_p=0.95) are provided in the prompt. Score must be clipped to [0,1]. Answer 0 if removing the segment would still leave the observed answer more likely than a different answer. Answer 1 if removing the segment would make a different answer more likely than the observed answer.

Return one floating-point score for each segment as JSON: \{"scores":\{"1":...,"2":...\}\}, with exactly one entry per candidate segment.

Original problem, observed final answer, observed full completion, and the candidate segment list are interpolated into the prompt as separate fields.
\end{promptbox}

\begin{promptbox}[title={Counterfactual sensitivity, variant v2 (GPT-5.4)}, label=box:cfvtwo]
Below is one observed run from an AI model on a problem. The problem text has been divided into candidate segments.

You will see the original problem, the model's observed full completion, the observed final answer, and the candidate segments.

For each candidate segment, imagine removing only that segment from the problem. Predict how much that removal would reduce the chance that the model gives the same final answer as in the observed run.

Technical details and score range match variant v3. Answer 0 if, after removing the segment, the same final answer is more likely than a different final answer; answer 1 if a different final answer is more likely.

Return one floating-point score per segment as JSON, with exactly one entry per candidate segment.
\end{promptbox}

\paragraph{Rerun consistency prompts.}
All three variants ask the reader to forecast how often the observed final answer would recur on reruns, differing only in how the score is framed.
GPT-5.4 and Claude Opus~4.6 both used variant (a), the direct stability prompt (Box~\ref{box:reruna}).
Variant (b) reframes the target as ``the probability that another run on the same input would produce the same final answer'' and variant (c) as ``the expected fraction of many reruns whose final answer would match the observed final answer''; both are included in the released prompt set.

\begin{promptbox}[title={Rerun consistency, variant (a) (both readers)}, label=box:reruna]
You are forecasting how stable a reasoning model's final answer is under rerunning.

You are given the exact input the model saw, the model's observed final answer on this run, and the model's full completion on this run.

Task: predict the stability score of this observed answer on a scale from 0 to 1. 0 means this answer would be unique among repeated runs on the same input; 1 means essentially all repeated runs would give this same answer; intermediate values mean this answer would recur on some but not all reruns.

Predict answer stability under rerunning, not correctness. Base your prediction only on the observed input and this observed run. Return only a single decimal number between 0 and 1, with no explanation.
\end{promptbox}

\paragraph{Behavior Forecaster and naive reader test metrics.}
Table~\ref{tab:appendix_naive_reader_metrics} reports the OLMo-3-7B-Think test Spearman and Pearson correlations summarized by the top row of Figure~\ref{fig:learnability_overview}.
Behavior Forecaster numbers are the test results used for the paper-facing comparison.

\begin{table}[t]
\centering
\small
\caption{OLMo-3-7B-Think test Spearman and Pearson correlation for the trained Behavior Forecaster and two frontier naive readers on each task.}
\label{tab:appendix_naive_reader_metrics}
\setlength{\tabcolsep}{8pt}
\begin{tabular}{l cc cc}
\toprule
& \multicolumn{2}{c}{\textbf{Counterfactual sensitivity}} & \multicolumn{2}{c}{\textbf{Rerun consistency}} \\
\cmidrule(lr){2-3} \cmidrule(lr){4-5}
& Spearman & Pearson & Spearman & Pearson \\
\midrule
Behavior Forecaster      & $\mathbf{0.731}$ & $\mathbf{0.931}$ & $\mathbf{0.571}$ & $\mathbf{0.667}$ \\
GPT-5.4 (naive)          & $0.672$          & $0.890$          & $0.305$          & $0.359$          \\
Claude Opus~4.6 (naive)  & $0.690$          & $0.888$          & $0.308$          & $0.537$          \\
\bottomrule
\end{tabular}
\end{table}

Table~\ref{tab:appendix_main_per_dataset} breaks the OLMo-3-7B-Think comparison down by dataset family.
The Behavior Forecaster is more accurate than both naive readers in aggregate and on most per-dataset cells.

\begin{table}[t]
\centering
\small
\caption{Per-dataset test Spearman/Pearson correlations for the OLMo-3-7B-Think comparison in Figure~\ref{fig:learnability_overview}. Behavior Forecaster values are computed from the last saved checkpoints: step $1668$ for counterfactual sensitivity and step $1336$ for rerun consistency. For counterfactual sensitivity, the paper-facing plotted checkpoint is step $1600$; the last saved checkpoint differs only in the fourth decimal place in the pooled metric. FEVEROUS aggregates the four FEVEROUS subsets, matching the three-dataset framing used in the main text.}
\label{tab:appendix_main_per_dataset}
\setlength{\tabcolsep}{5pt}
\resizebox{\linewidth}{!}{%
\begin{tabular}{llrcccc}
\toprule
\textbf{Task} & \textbf{Dataset} & \textbf{$n$} & \textbf{Behavior Forecaster} & \textbf{Claude Opus~4.6} & \textbf{GPT-5.4} & \textbf{Single-location probe} \\
\midrule
\multirow{3}{*}{Counterfactual sensitivity}
& FEVEROUS & $3{,}155$ & $0.660/0.775$ & $0.574/0.578$ & $0.621/0.674$ & $0.209/0.116$ \\
& RuleTaker & $11{,}740$ & $0.668/0.967$ & $0.649/0.961$ & $0.601/0.949$ & $-0.008/0.017$ \\
& TreeCut & $1{,}425$ & $0.840/0.946$ & $0.775/0.865$ & $0.804/0.887$ & $0.258/0.186$ \\
\midrule
\multirow{3}{*}{Rerun consistency}
& FEVEROUS & $375$ & $0.580/0.517$ & $0.607/0.475$ & $0.537/0.370$ & $0.495/0.489$ \\
& RuleTaker & $375$ & $0.627/0.887$ & $0.069/0.692$ & $0.245/0.380$ & $0.057/-0.012$ \\
& TreeCut & $375$ & $0.140/0.455$ & $-0.043/0.365$ & $0.058/0.248$ & $0.098/0.095$ \\
\bottomrule
\end{tabular}
}
\end{table}

\subsection{Single-location probe baselines on the target LRM}
\label{sec:appendix_single_location_probes}

The dashed single-location probes in Figure~\ref{fig:learnability_overview} are direct adaptations of standard attribution and uncertainty signals to each task: each reads a single fixed location of the frozen target LRM and produces a per-instance score, with no further training, threshold tuning, or calibration.
They are computed on the same test surfaces as the main figure.

\paragraph{Answer likelihood for rerun consistency.}
For each observed trajectory $\tau_i=(P_i,R_i,A_i)$, we teacher-force the stored prompt and completion through OLMo-3-7B-Think and score only the final-answer tokens after the first closing \texttt{</think>} tag.
The reported score is the mean final-answer log probability,
\[
s_i^{\mathrm{lik}} =
\frac{1}{|T(A_i)|}
\sum_{t \in T(A_i)}
\log p_M(x_t \mid x_{<t}),
\]
where $T(A_i)$ is the set of final-answer token positions.
Higher values mean lower answer perplexity and therefore higher predicted rerun consistency.
This baseline follows the broad use of token probabilities as uncertainty or hallucination signals in generation baselines~\citep{manakul2023selfcheckgpt,farquhar2024semanticentropy}; here we intentionally use the simplest answer-only version rather than semantic clustering or resampling.

\paragraph{Answer-to-input attention for counterfactual sensitivity.}
For each labeled input segment, we teacher-force the observed trajectory through OLMo-3-7B-Think with attention outputs enabled.
We then use the final-answer tokens as queries and score a candidate segment by the last-layer attention mass assigned to that segment's prompt tokens:
\[
s_j^{\mathrm{attn}} =
\frac{1}{H|T(A)|}
\sum_{h=1}^{H}
\sum_{t \in T(A)}
\sum_{u \in T(S_j)}
\alpha^{(L,h)}_{t,u}.
\]
Here $T(S_j)$ is the token span of segment $j$, $\alpha^{(L,h)}$ is head $h$ in the final transformer layer, and higher values mean that final-answer tokens attend more to the segment.
This sits in the standard family of attention-based attribution probes for context use~\citep{chuang2024lookbacklens, jain2019attention}, applied here at the natural readout point in an LRM: from the final-answer tokens back to the prompt.

\begin{center}
\small
\refstepcounter{table}
\label{tab:appendix_single_location_probes}
\textbf{Table~\thetable:} Single-location probe baselines on the target LRM, used as dashed lines in Figure~\ref{fig:learnability_overview}. Both are evaluated on the same test surfaces as the main figure.

\vspace{0.5em}
\setlength{\tabcolsep}{8pt}
\begin{tabular}{l l c c c}
\toprule
\textbf{Task} & \textbf{Probe} & \textbf{$n$} & \textbf{Spearman} & \textbf{Pearson} \\
\midrule
Counterfactual sensitivity & Answer-to-input attention mass & $16{,}320$ & $0.160$ & $0.160$ \\
Rerun consistency & Final-answer likelihood & $1{,}125$ & $0.129$ & $0.118$ \\
\bottomrule
\end{tabular}
\end{center}

\subsection{Target-LRM naive reader control}
\label{sec:appendix_target_model_naive_reader}

As an additional diagnostic, we evaluate OLMo-3-7B-Think itself as a naive reader of its own observed trajectory.
This baseline tests whether the target LRM can recover the future-behavior signal by prompted reading alone, without any training or calibration.
In the stateless variant, one prompt contains the original input, the observed OLMo completion, the observed final answer, and the same forecasting task given to the other naive readers.
In the stateful variant, we replay the original input as the user message and the observed OLMo completion as the assistant message, then append a follow-up user message asking for the forecast.

\begin{table}[t]
\centering
\small
\caption{Appendix-only target-LRM naive reader control. OLMo-3-7B-Think is prompted to read its own observed trajectory and predict the target behavioral property without training.}
\label{tab:appendix_target_model_naive_reader_metrics}
\setlength{\tabcolsep}{8pt}
\begin{tabular}{l cc cc}
\toprule
& \multicolumn{2}{c}{\textbf{Counterfactual sensitivity}} & \multicolumn{2}{c}{\textbf{Rerun consistency}} \\
\cmidrule(lr){2-3} \cmidrule(lr){4-5}
& Spearman & Pearson & Spearman & Pearson \\
\midrule
OLMo as naive reader (stateless) & $0.167$ & $0.182$ & $-0.025$ & $0.019$ \\
OLMo as naive reader (stateful)  & $0.226$ & $0.245$ & $0.159$  & $0.197$ \\
\bottomrule
\end{tabular}
\end{table}

\section{Statistical uncertainty estimates}
\label{sec:appendix_statistical_uncertainty}

For the main learnability comparison on OLMo-3-7B-Think, we estimate uncertainty by cluster bootstrapping over original samples, stratified by dataset family.
This treats an original sample, not a flattened segment or observed run, as the resampling unit.
We use $5{,}000$ bootstrap resamples for confidence intervals.
For paired tests, we compare the Behavior Forecaster and each naive reader on exactly matched prediction rows, resampling the same sample clusters for both systems.
We also run a one-sided paired permutation test with $10{,}000$ permutations, swapping the two systems' predictions together at the original-sample cluster level and using the directional alternative that the Behavior Forecaster has higher correlation than the naive reader.
Table~\ref{tab:appendix_main_uncertainty} reports Spearman uncertainty for the main metric.

\begin{table}[t]
\centering
\small
\caption{Cluster-bootstrap uncertainty and one-sided paired permutation tests for the main Spearman results. CIs are $95\%$ cluster-bootstrap intervals over original samples.}
\label{tab:appendix_main_uncertainty}
\setlength{\tabcolsep}{4pt}
\resizebox{\linewidth}{!}{%
\begin{tabular}{l l c c c c}
\toprule
\textbf{Task} & \textbf{Reader} & \textbf{Forecaster} & \textbf{Reader} & \textbf{Paired gap} & \textbf{One-sided perm. $p$} \\
\midrule
Counterfactual sensitivity & GPT-5.4
& $0.731$ [$0.701$, $0.756$]
& $0.672$ [$0.629$, $0.713$]
& $0.059$ [$0.027$, $0.088$]
& $0.011$ \\
Counterfactual sensitivity & Claude Opus~4.6
& $0.731$ [$0.701$, $0.756$]
& $0.690$ [$0.650$, $0.727$]
& $0.041$ [$0.013$, $0.068$]
& $0.037$ \\
Rerun consistency & GPT-5.4
& $0.571$ [$0.476$, $0.657$]
& $0.305$ [$0.208$, $0.398$]
& $0.266$ [$0.162$, $0.361$]
& $<0.001$ \\
Rerun consistency & Claude Opus~4.6
& $0.571$ [$0.476$, $0.657$]
& $0.308$ [$0.196$, $0.412$]
& $0.263$ [$0.167$, $0.361$]
& $<0.001$ \\
\bottomrule
\end{tabular}
}
\end{table}

For transfer, the unit of replication is the held-out target dataset, so the analysis is necessarily low-power with three targets per task.
The adapted source improves over the unadapted source for all three held-out targets on both tasks.
The mean adapted-minus-source Spearman gap is $0.228$ with bootstrap CI [$0.196$, $0.273$] for counterfactual sensitivity and $0.268$ with CI [$0.075$, $0.433$] for rerun consistency; the one-sided sign-test value is $p=0.125$ in both cases because there are only three targets.
We treat ablation rows as descriptive single-run comparisons rather than significance tests, since they do not have repeated seeds or saved paired predictions for every variant.

\end{document}